\documentclass{article}

\usepackage{arxiv}

\usepackage[utf8]{inputenc} 
\usepackage[T1]{fontenc}    
\usepackage{hyperref}       
\usepackage{url}            
\usepackage{booktabs}       
\usepackage{amsfonts}       
\usepackage{nicefrac}       
\usepackage{microtype}      
\usepackage{lipsum}		
\usepackage{graphicx}
\usepackage{natbib}
\usepackage{doi}
\usepackage{alterqcm}
\usepackage[dvipsnames]{xcolor}

\title{Spanish and LLM Benchmarks: is MMLU Lost in Translation?    }

\author{ Irene Plaza, Nina Melero, Cristina del Pozo, Javier Conde and Pedro Reviriego  \\
	ETSI de Telecomunicación\\
	Universidad Politécnica de Madrid\\
	28040 Madrid, Spain \\
	\And
        {Marina Mayor-Rocher} \\
        Facultad de Filosofía y Letras \\
	Universidad Autónoma de Madrid\\
        28049 Madrid, Spain \\
        \And
        {María Grandury} \\
        SomosNLP \\
	24402, Ponferrada, Spain\\
}

\renewcommand{\shorttitle}{\textit{arXiv} Template}

\begin{document}
\maketitle

\begin{abstract}

The evaluation of Large Language Models (LLMs) is a key element in their continuous improvement process and many benchmarks have been developed to assess the performance of LLMs in different tasks and topics. As LLMs become adopted worldwide, evaluating them in languages other than English is increasingly important. However, most LLM benchmarks are simply translated using an automated tool and then run in the target language. This means that the results depend not only on the LLM performance in that language but also on the quality of the translation. In this paper, we consider the case of the well-known Massive Multitask Language Understanding (MMLU) benchmark. Selected categories of the benchmark are translated into Spanish using Azure Translator and ChatGPT4 and run on ChatGPT4. Next, the results are processed to identify the test items that produce different answers in Spanish and English. Those are then analyzed manually to understand if the automatic translation caused the change. The results show that a significant fraction of the failing items can be attributed to mistakes in the translation of the benchmark. These results make a strong case for improving benchmarks in languages other than English by at least revising the translations of the items and preferably by adapting the tests to the target language by experts. 

\end{abstract}


\section{Introduction}
\label{sec:intro}

Large Language Models are becoming a fundamental block in modern computing systems enabling new applications and facilitating the interaction with users \cite{LLM_survey}. However, LLMs have limitations and their performance has to be well understood before using them on a given application \cite{LLM_survey2}. This has motivated the development of a large number of LLM evaluation benchmarks that test the knowledge that models have of many different topics and how well they can perform tasks such as logic reasoning or problem-solving \cite{LLM_evaluation_survey}. Most of these benchmarks are designed so that the LLM responses can be processed automatically thus enabling testing at scale with thousands of questions. This is commonly achieved by using multiple-choice tests. 

There are LLM benchmarks to evaluate a wide range of tasks and topics. For example, there are tests to evaluate the capabilities of LLMs to solve common sense reasoning problems \cite{zellers2019hellaswag} or to answer mathematical questions \cite{Mathmeasuring}. To provide a more comprehensive evaluation, some benchmarks evaluate several tasks, for example, the Multitask Language Understanding (MMLU) test \cite{MMLU} evaluates 57 different topics and other benchmarks increase the number of tasks and topics to more than 200 \cite{BIGMeasuring}. The speed and energy dissipation of LLMs are also important factors that are commonly evaluated in terms of the number of tokens generated per second, the memory used or the energy per token\footnote{\url{https://huggingface.co/spaces/optimum/llm-perf-leaderboard}}, or with more user-centric metrics like the time and energy needed to complete a given task \cite{SpeedLLM}. As LLMs become pervasive and used in almost any domain and application, more benchmarks will be developed each having possibly thousands or even hundreds of thousands of questions.      

Another dimension of LLM evaluation is their performance in languages other than English as in fact most users are native speakers of other languages. Most benchmarks are written in English with questions taken in many cases from different exams, such as university, high school or professional tests. The simplest approach is to translate these same tests into other languages and use them for multilingual evaluation. This clearly introduces a cultural bias, especially when questions are related to subjects such as history, geography, art or general culture. Ideally, specific tests should be developed or at least adapted for each language.

However, this is not the only problem. To be able to evaluate LLMs in many languages, and given the large number of questions of the benchmarks, the standard procedure is to translate the English test to the target language using automatic translation tools, for example, in the Okapi project \cite{lai2023okapi} three benchmarks from the Open LLM Leaderboard \cite{open-llm-leaderboard} are translated using ChatGPT while in the evaluation of GPT4, the tests were translated using Azure Translator \cite{GPT4}. This implies that the benchmarks in languages other than English are not only measuring the performance of the LLM but also of the translation tool as the quality of the translation can clearly impact the results. 

In this work we perform an initial analysis of the impact of automatic translation on one of the most widely used LLM benchmarks, the Multitask Language Understanding (MMLU) test \cite{MMLU} for one of the most commonly used and chosen as a second language to learn, Spanish. The analysis shows that automatic translation induces errors in the LLM answers and thus distorts the benchmark's results. Based on these findings, potential solutions to this problem are also briefly discussed.

The rest of the work is organized as follows, in section \ref{sec:methodology} the methodology used in our analysis is presented, followed by the results in section \ref{sec:ResultsAnalysis} and a discussion of their implications and potential solutions in section \ref{sec:discussion}. The paper ends with the conclusion in section \ref{sec:Conclusion}.

\section{Methodology}
\label{sec:methodology}

This section discusses the methodology used in our analysis, first, the tests used in the evaluation and tools selected are discussed to then describe the evaluation procedure.

\subsection{Tests and Tools}

To evaluate the impact of automatic translation on the benchmarks, we have selected three categories from the MMLU test \cite{MMLU}: Miscellaneous, Philosophy, and US foreign policy with 783, 311, and 100 questions respectively. The first one covers a wide range of topics that can be affected by translation, while the second one focuses on well-known content that is universal. Finally, the last category focuses on US-related questions that may also be prone to translation errors. Therefore, the three categories can provide insights into the potential limitations of automatic translation.

In terms of translation tools, we consider two: Azure Translator and ChatGPT. The reasoning behind our choices is that Azure Translator was the tool used to evaluate MMLU for languages other than English in GPT4 \cite{GPT4}. Therefore, our findings would be applicable to GPT4 evaluation results. On the other hand, it is interesting to check how well an LLM performs when translating tests that will be used to evaluate the same LLM. Thus we translate the questions with ChatGPT4 and then use them to evaluate it. This enables us to check whether using the same tool for translation and testing introduces any bias in the evaluation, for example, better performance may be achieved when the same tool is used for both.

Finally, the LLM used to answer the questions, both in English and Spanish, is GPT4, which is among the best-performing LLMs on the MMLU benchmark to date. The rationale is that this model will produce the largest number of correct answers in the absence of translation errors and thus will be the most effective in detecting discrepancies due to translation errors. For example, a model that already produces wrong answers for 50\% of the questions in English will not detect translation errors on that 50\%. In the same way, it is possible that right answers in English are incorrectly translated into Spanish but answered correctly by the LLM. 

\subsection{Evaluation Procedure}

\begin{figure*}[b]
  \centering
  \includegraphics[scale=0.4]{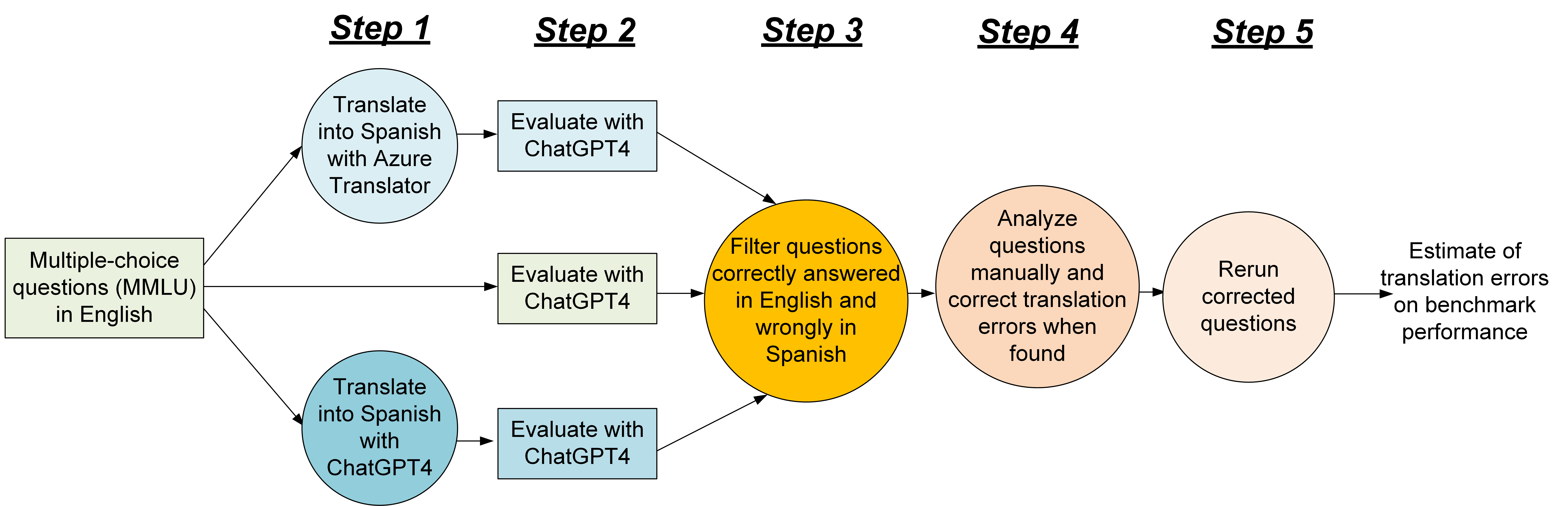}
  \caption{Diagram of the evaluation methodology used.}
  \label{fig:Methodology}
\end{figure*}

The overall procedure is illustrated in Figure \ref{fig:Methodology} and has the following steps:

\begin{enumerate}
    \item Translate the questions from the selected categories into Spanish using the chosen tools: Azure Translator and ChatGPT4.
    \item Run the same MMLU tests on both the original and the Spanish translated versions and log the answers and their log probabilities.
    \item Identify the questions in which the answers are correct in English and wrong in Spanish.
    \item Analyze them manually. Two experts revise each question independently to avoid mistakes or biases in the evaluation and the questions are divided into two groups: correct and having translation errors. Revise and correct the translation errors.
    \item Run the corrected questions again on the LLM. The number of correct answers is logged and reported to estimate the impact of translation errors on the benchmark results. 
\end{enumerate}

The procedure is designed to evaluate the impact of the errors in the automatic translation on the test results. This enables us to focus only on the questions that are answered correctly in English and wrongly in any of the two Spanish translations, significantly reducing the effort needed for the manual check of the translation. This is relevant as it enables the methodology to scale to larger datasets with a reasonable effort and cost as each question is checked and corrected by two experts.   
 
\subsection{Limitations}

The proposed methodology has a number of limitations. Firstly, it cannot detect errors in translation that do not affect the answer, i.e., if there is a mistake in a translation but the LLM answer is correct it will not be detected. This can be due to the translation error not affecting the LLM's ability to answer but also due to randomness in the responses. Secondly, the fact that the LLM answers correctly to the revised translation may also be due to randomness in some cases. 

Finally, there are also limitations related to the number of questions tested and the tools used. The evaluation was limited to approximately 1200 questions and only two automatic translation tools to keep the effort of manual revisions acceptable. To get a better estimate of the impact of automatic translations, more questions taken from different benchmarks and using several translation tools should be used. The results also depend on the prompts used to ask the LLM, the LLM used, and its configuration parameters. 

\section{Results and Analysis}
\label{sec:ResultsAnalysis}

As per the proposed methodology, first, the questions are translated into Spanish with Azure Translator and ChatGPT. The translations as well as the rest of the results presented in the following, are available in a public repository\footnote{https://zenodo.org/records/11314109}. Next, the three versions of the selected MMLU questions are run on ChatGPT4 and the results in terms of the number and percentage of incorrect answers are summarized in Table \ref{tab:table1}. It can be observed that there is a performance loss when the tests are run in Spanish in all three categories and for both Azure and ChatGPT translations. The relative loss is significantly larger in Philosophy and US foreign policy. 

\begin{table}
	\caption{Number (percentage) of incorrect answers for ChatGPT4.}
	\centering
	\begin{tabular}{llll}
		Category     & English     & Spanish (Azure) & Spanish (ChatGPT) \\
		\midrule
		Miscellaneous & 47 (6\%)  & 65  (8.3\%) & 66 (8.56\%)     \\
		  Philosophy & 43 (13.9\%)   & 66 (21.29\%) &  57 (18.4\%)    \\
		US foreign policy & 3 (3\%) & 10 (10\%)  & 11 (11\%)      \\
		\bottomrule
	\end{tabular}
	\label{tab:table1}
\end{table}

The third step is to identify the questions that were correctly answered in English but failed in Spanish.
The number of such questions is given in Table \ref{tab:table2}. For US foreign policy, adding those to the failures in English gives the number of failures in Spanish. This means that there are no correct answers in Spanish for items that failed in English. Instead, for the other two categories, the addition is larger than the number of failures in Spanish, which means that there are a few correct answers in Spanish for items that failed in English or that failed in both languages. Comparing the translations, the one done with Azure Translator has more questions correctly answered in English but failed in Spanish in total, and in two of the three categories. This may be due to ChatGPT understanding better its own translations.

\begin{table}
	\caption{Number of correct answers in English and wrong in Spanish for ChatGPT4.}
	\centering
	\begin{tabular}{lll}
		Category         & Spanish (Azure) & Spanish (ChatGPT) \\
		\midrule
		Miscellaneous    &  33 &  28                           \\
            Philosophy   & 29 &  21                             \\
		US foreign policy  & 7 & 8                               \\
		\bottomrule
	\end{tabular}
	\label{tab:table2}
\end{table}

In the fourth step, the questions with correct answers in English and wrong in Spanish were analyzed manually to check the translations done by Azure Translator and ChatGPT. There are clear examples of errors which can be attributed to the translation. In the Miscellaneous category in the question below, the correct answer is John Constable. For some reason, Azure Translator translates the name to ``Juan Alguacil'' (in boldface) but not the others and ChatGPT4 selects an incorrect answer. Instead, the answer is correct for the ChatGPT translation as it does not translate John Constable.    

\begin{alterqcm}[lq=8cm,language=english, correction, corsymb=\dingchecksquare] 
 \AQquestion[br={1}]{Original question: \\ Which of the following artists is an English landscape painter who created a poetic sense of changing atmospheric effects by using tiny applications of local colour? Please answer:}{%
 {\textbf{\textcolor{OliveGreen}{John Constable}}},
 {William Morris},
 {William Hogarth},
 {Joshua Reynolds}}  
\AQquestion[br=1]{ChatGPT translated question: \\ ¿Cuál de los siguientes artistas es un pintor paisajista inglés que creó un sentido poético de los cambios en los efectos atmosféricos utilizando pequeñas aplicaciones de color local? Por favor responda:}{%
 {\textbf{\textcolor{OliveGreen}{John Constable}}},
 {William Morris},
 {William Hogarth},
 {Joshua Reynolds}}  
\AQquestion[br=4]{Azure Translator question: \\ ¿Cuál de los siguientes artistas es un pintor paisajista inglés que creó un sentido poético de los efectos atmosféricos cambiantes mediante el uso de pequeñas aplicaciones de color local? Por favor, responda:}{%
 {\textbf{\textcolor{OliveGreen}{Juan Alguacil}}},
 {William Morris},
 {William Hogarth},
 {\textbf{\textcolor{Red}{Joshua Reynolds}}}}  
\end{alterqcm}

Similarly in the US foreign policy category, we have the following question that is wrongly answered in Spanish for both Azure Translator and ChatGPT translations. In this case, the issue is that the expression ``American multiplication table'' is translated literally into Spanish and loses its meaning of the large population growth in the US \cite{sexton2018nation}. Actually, there is no translation for this expression, which points to a fundamental limitation of translating benchmarks: there may be questions that cannot be translated.    
 
\begin{alterqcm}[lq=8cm,language=english, correction, corsymb=\dingchecksquare] 
 \AQquestion[br=1]{Original question: \\ What was meant by the term ``American multiplication table''? Please answer:}{%
 {\textbf{\textcolor{OliveGreen}{Increase in the US population}}},
 {Increase in US finances},
 {Increase in US military capability},
 {Increase in US international influence}}  
\AQquestion[br=4]{ChatGPT translated question: \\  ¿Qué se entendía por el término ``tabla de multiplicación americana''? Por favor responda:}{%
  {\textbf{\textcolor{OliveGreen}{Aumento en la población de EE.UU}}},
 {Aumento de las finanzas de EE.UU},
 {Aumento en la capacidad militar de EE.UU},
 {\textbf{\textcolor{Red}{Aumento en la influencia internacional de EE.UU}}}}  
\AQquestion[br=4]{Azure Translator question:\\  ¿Qué se entendía por el término ``tabla de multiplicación americana''?  Por favor, responda:}{%
  {\textbf{\textcolor{OliveGreen}{Aumento en la población de EE.UU}}},
 {Aumento de las finanzas de EE.UU},
 {Aumento en la capacidad militar de EE.UU},
 {\textbf{\textcolor{Red}{Aumento de la influencia internacional de Estados Unidos}}}}
\end{alterqcm}

\clearpage
In addition to checking the translations to identify and correct mistakes, an analysis of those mistakes was made to try to find patterns in the errors. The conclusion was that the translation errors observed can be grouped into the following main categories:

\begin{enumerate}
    \item Translation of a proper name: ``John Constable'' translated into ``Juan Alguacil'', or ``Stephen King'' as “Esteban Rey” which seems to demonstrate that the model does not identify proper names as entities.
    \item Incorrect translation of a technical term: ``Furrow opener'' translated into ``abre surcos'' instead of ``arado'', ``lieutenant general'' as ``general de teniente'' instead of ``teniente general''.
    \item A term is not translated: ``Wednesday child'' translated into ``El hijo de Wednesday'', ``Cicero'' not translated into ``Cicerón''.
    \item Cultural adaptation was needed in the translation: ``The paper chase'' translated into ``La persecución del papel'', which is not consistent with the name used in Spain; and measures (length, currencies) are not adapted.
    \item Change of meaning:  "grand” is incorrectly translated into Spanish as ``big'', ``older'' as ``old'' instead of as ``great'', ``rules make take into account...'' as ``rules may take into account...''. 
    \item Grammatical errors: the words in the question are feminine and the ones used in the options appear as masculine or vice versa; and the text sometimes contains ungrammatical sequences caused by literal translations (``how many pence make a pound?'' translated as ``¿cuántos peniques hacen una libra?''. 
\end{enumerate}

Table \ref{tab:table3} shows the results of the analysis done on the questions that had correct answers in English and wrong in Spanish with the initial translation. First, the number of questions in which the manual checking found issues in the automatic translation is given. It can be observed that for Azure translator the majority of the questions have errors in the translation while for ChatGPT the numbers are lower but still significant. 

As per the proposed methodology, in step 5, the incorrectly translated questions were translated manually and rerun on ChatGPT4. The results in terms of questions that are incorrectly answered after fixing the translation are also shown in \ref{tab:table3}. It can be seen that a significant fraction of the questions have the right answer after manually fixing the translation. In the case of the Azure translations, the percentage is above 34\% for all three categories and exceeds 63\% for Miscellaneous. In the case of ChatGPT, the number of questions corrected is again lower in total and lower in two of the three categories with values ranging from 9\% to 37\%. These results clearly illustrate how the use of automatic translations can impact the results of benchmarks ran in languages other than English.   

\begin{table}
	\caption{Analysis of the questions that had correct answers in English and wrong in Spanish for ChatGPT4.}
	\centering
	\begin{tabular}{llllll}
		Category    &  Translation    & Initial & Translation issues & After translation revision & Percentage corrected  \\
		\midrule
		Miscellaneous    &  Azure & 33 & 29 &  12  & 63.64\%                    \\
		Miscellaneous    &  ChatGPT & 28 & 15  &  17  &   32.14\%                 \\
		Philosophy    &  Azure & 29 & 25 &  19  & 34.48\%                    \\
		Philosophy    &  ChatGPT & 21 & 4  & 19   &   9.52\%                 \\
		US foreign policy    &  Azure & 7 & 5 &  4  & 42.86\%                    \\
		US foreign policy    &  ChatGPT & 8 & 6 &  5  &   37.5\%                 \\
		\bottomrule
	\end{tabular}
	\label{tab:table3}
\end{table}

\section{Discussion}
\label{sec:discussion}

The findings of our initial evaluation suggest that the use of automatic translation of the questions in LLM benchmarks causes deviations in the 
results due to errors in the translation. To eliminate those deviations and ensure that the results are not contaminated by translation errors, the translations should be revised by experts and, ideally, the tests should be adapted to the language being evaluated. However, given the number of questions on the LLM benchmarks and the number of languages evaluated, this requires a large effort that calls for coordinated action from the community. 

In fact, there are efforts in this direction such as the validation of the Spanish Okapi benchmarks as part of the \#Somos600M Project \cite{grandury2024somos600m}. This community annotation effort uses open-source frameworks and is performed by native Spanish-speaking volunteers. During the first two months, more than 60 persons participated and together covered one-third of the total number of dataset items,
which shows how time-consuming it is to manually validate and correct these translations.

In the absence of manually adapted or checked benchmarks, our methodology enables a fast evaluation of the impact of translation errors with limited manual checking. Further refinements can be introduced to detect translation errors. For example, after translating into Spanish we could translate back into English the questions with different answers in both languages and run the questions again. When the answer to this English translation is different from the one of the original English question, the translation is likely to be the culprit of the error. In general, developing strategies to identify these issues automatically would be very helpful to understand the impact of automatic translation and also to correct the translation errors.

In this work, we have focused on Spanish which is one of the most widely used languages in the world and also is typically in the top five languages with more data on the LLM training datasets. It would be interesting to study the impact of translation errors on other languages. For languages with fewer data and speakers, we would expect a larger number of translation errors but also a larger number of genuine errors in the LLM answers. Therefore, the relative impact of translation errors on the benchmark results compared to Spanish can be either larger or smaller.

\section{Conclusion and future work} 
\label{sec:Conclusion}

In this work, we have analyzed the limitations of using automatic translation of English benchmarks to evaluate LLMs in other languages. In more detail, three categories from the MMLU benchmark have been translated into Spanish and run on ChatGPT4. As previously mentioned, it would be interesting to test other categories from the MMLU benchmark and other models. Then, the test items for which the LLM answers are different in English and Spanish have been identified and analyzed manually to understand if the differences can be attributed to the translation. The results show that a significant fraction of the differences are due to errors in the translation of the questions. These findings highlight the need to improve non-English LLM benchmarks by at least ensuring that the translations are correct and ideally by adapting the questions to the target language and culture.

The development of language-specific or at least language-adapted benchmarks should be a priority to provide better evaluation tools for multilingual LLMs. To achieve that goal, open initiatives are needed to coordinate the efforts of the community to develop for example language specific LLM leaderboards.

\bibliographystyle{apalike}
\bibliography{references}  

\end{document}